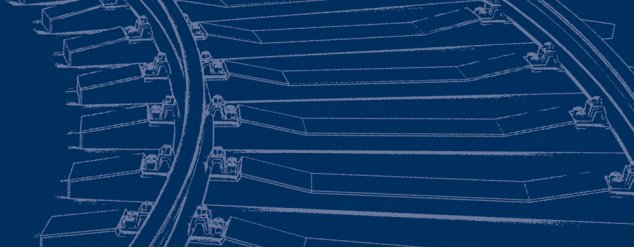

# Advanced technology in railway track monitoring using the GPR Technique: A Review

Farhad Kooban[1], Aleksandra Radlińska[1], Reza Mousapour[2], & Maryam Saraei[3]
[1]Department of Civil and Environmental Engineering, Penn State University, Pennsylvania, USA
[2]Department of Civil and Environmental Engineering, University of Alberta, Edmonton, Alberta, Canada
[3]Department of Railway Engineering, Iran University of Science and Technology, Tehran, Iran

ABSTRACT

Subsurface evaluation of railway tracks is crucial for safe operation, as it allows for the early detection and remediation of potential structural weaknesses or defects that could lead to accidents or derailments. Ground Penetrating Radar (GPR) is an electromagnetic survey technique as advanced non-destructive technology (NDT) that can be used to monitor railway tracks. This technology is well-suited for railway applications due to the sub-layered composition of the track, which includes ties, ballast, sub-ballast, and subgrade regions. It can detect defects such as ballast pockets, fouled ballast, poor drainage, and subgrade settlement. The paper reviews recent works on advanced technology and interpretations of GPR data collected for different layers. Further, this paper demonstrates the current techniques for using synthetic modeling to calibrate real-world GPR data, enhancing accuracy in identifying subsurface features like ballast conditions and structural anomalies and applying various algorithms to refine GPR data analysis. These include Support Vector Machine (SVM) for classifying railway ballast types, Fuzzy C-means, and Generalized Regression Neural Networks for high-accuracy defect classification. Deep learning techniques, particularly Convolutional Neural Networks (CNNs) and Recurrent Neural Networks (RNNs) are also highlighted for their effectiveness in recognizing patterns associated with defects in GPR images. The article specifically focuses on the development of a Convolutional Recurrent Neural Network (CRNN) model, which combines CNN and RNN architectures for efficient processing of GPR data. This model demonstrates enhanced detection capabilities and faster processing compared to traditional object detection models like Faster R-CNN.

## 1 INTRODUCTION

Rail transportation is a compelling alternative to road transport, offering superior efficiency in key areas like energy consumption, CO2 emissions, system capacity, and safety (Alexander 2012). However, railway tracks are susceptible to internal defects that can worsen over time, leading to significant deterioration. This deterioration includes issues ballast fouling voids beneath sleepers, washouts, subgrade diseases (Liu et al. 2023), moisture levels in ballast and ballast beds, water cavities due to drainage problems, and locations with insufficient bearing capacity (e.g., ballast pockets). These issues can compromise track safety and performance. To address these concerns, conducting regular and thorough inspections to monitor the internal condition of tracks is crucial. Given that these problems often form within the underground structure, which is invisible to the naked eye, current inspection methods rely on observing surface manifestations, such as rail corrugation (Esmaeili 2014), white spots which is like limestone powder on tracks (Sysyn, et al. 2020 & Mostofinejad et al. 2021) or mud pumping (Figure 1). In contrast, non-destructive testing methods offer viable alternatives. These methods enable the evaluation of subsurface conditions without causing harm to the structure, making them safer and more practical for routine inspections. A variety of non-destructive testing techniques are available for assessing the subsurface condition of transportation infrastructure, including analyzing vibration trends and detecting abnormal vibrations (Sui et al. 2023).

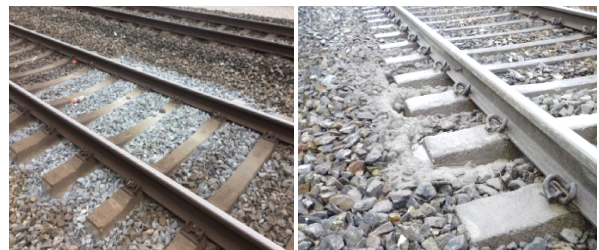

Figure 1. White spots of ballast breakdown (left) (Sysyn, et al. 2020) and Mud pumping induced by ballast fouling (right) (Nurmikolu et. Al (2013)

However, these techniques sometimes lack the accuracy needed to identify problems beneath the track structure.

Among various non-destructive testing methods, Ground Penetrating Radar (GPR) is particularly effective for railway track inspections (Solla et al. 2021). First used in the 1990s for the detection of railway infrastructure issues, GPR initially employed ground-mounted antennas. Subsequent advancements led to the adoption of high-frequency horn antennas, which facilitate non-contact testing and enable faster data acquisition (Liu et al. 2023). GPR is an electromagnetic method that images the subsurface using radar pulses. It operates by sending a high-frequency electromagnetic signal into the ground via a transmitting antenna. When this signal encounters different materials or interfaces within the subsurface, such as between ballast, sub-ballast, and the subgrade (Feld 2017; Casas et al. 2009), part of the signal is reflected back to the surface and captured by a receiving antenna (Figure 2). The time taken for the signal to return, and the strength of the reflected signal provide information about the depth and nature of the subsurface features (Al-Qadi et al. 2008).

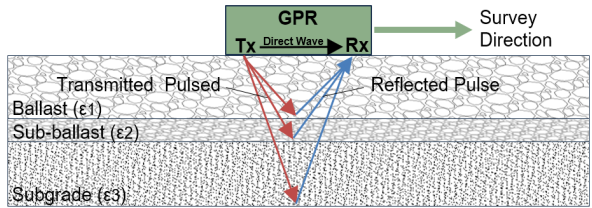

Figure 2. Schematic diagram of a GPR survey shows how reflected waves, as they move beneath the surface, travel at different velocities depending on the dielectric permittivity (ε) of the materials they come across

GPR's effectiveness in railway track inspection stems from its ability to detect variations in material composition, moisture content, and structural anomalies. This makes it invaluable for identifying common railway track defects such as ballast fouling, voids beneath sleepers, washouts, subgrade diseases (Liu et al. 2023), moisture levels in ballast and ballast beds, water cavities due to drainage problems, and locations with insufficient bearing capacity (e.g., ballast pockets). However, challenges arise in interpreting GPR data in high-conductivity materials (e.g., clay soils) and heterogeneous conditions (e.g., rocky beds).

The presence of wet beds or elevated moisture levels in railway track beds, especially in cold regions, can significantly impact railway infrastructure and safety. Excessive moisture in the subgrade can cause various issues, including subsidence and mud pumping (Cheng et al. 2022). Moreover, moisture within the track bed can reduce the subgrade's bearing capacity. When the ground becomes saturated, its strength and stiffness are compromised, potentially leading to track settling or shifting. This issue is particularly acute in areas with clayey soils, which are more prone to moisture content fluctuations.

Early detection through GPR provides crucial data for the maintenance and repair of the subgrade, offering valuable insights into its condition and informing necessary interventions (Cheng et al. 2022). The vehicle mounted GPR device enables rapid and efficient non-destructive inspections of the subgrade's data without interrupting train operations, as shown in Figure 3.

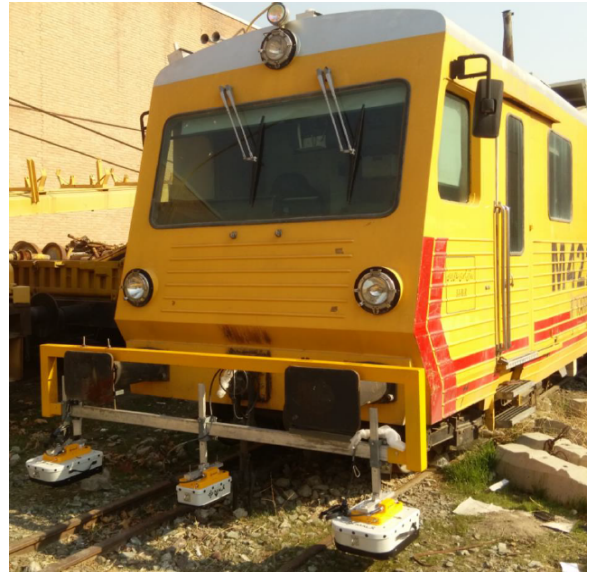

Figure 3. Data collection setup of two 500 MHz antennas and one 800 MHz antenna for GPR test on railway track beds

## 2 A REVIEW OF GPR PRINCIPLES

Ground penetrating radar (GPR) is an imaging technique that operates by measuring reflected electromagnetic (EM) waves (Annan 2009; Travassos et al. 2017). It utilizes radar pulses in the ultra or very high frequencies (UHF/VHF) of the microwave portion of the radio spectrum.

The GPR system includes a transmitting di-pole antenna that emits EM pulses into the ground and a receiving di-pole antenna that records changes in the reflected signal over time. These reflections primarily occur at interfaces between materials with differing dielectric properties. The thickness of a surveyed layer is determined by using the time difference measured between the reflected pulses at its boundaries. This measurement is combined with the layer's known dielectric properties, as outlined in Equation 1 (Kahil et al. 2023):

$$d_i = \frac{c \times t_i}{2\sqrt{\varepsilon_{r,i}}} \quad [1]$$

where $d_i$ the thickness of a surveyed layer, $t_i$ the two-way reflection time, $c$ the speed of light and $\varepsilon_{r,i}$ the relative dielectric constant of the medium.

It is crucial to understand that even minor variations in dielectric impedance, such as small-scale heterogeneities, can result in faint or barely noticeable responses in GPR readings. These variations significantly impact the signal's energy as it propagates, primarily due to scattering phenomena. Notably, the extent of these scattering losses is closely tied to the frequency of the radar pulses; the losses increase with higher pulse frequencies (O'Neill, K., 1999; De Bold 2015). These interface reflections produce responses that are instrumental in deducing underground structural profiles.

This principle is exemplified in Figure *4*, which juxtaposes a railway profile (depicted on the left) against a typical radar line scan response profile (shown in the middle). When multiple scans are combined, they create an integrated underground radar profile (displayed on the right). The layered analysis capability of GPR makes it exceptionally suitable for railway applications, offering the advantage of high-speed data collection.

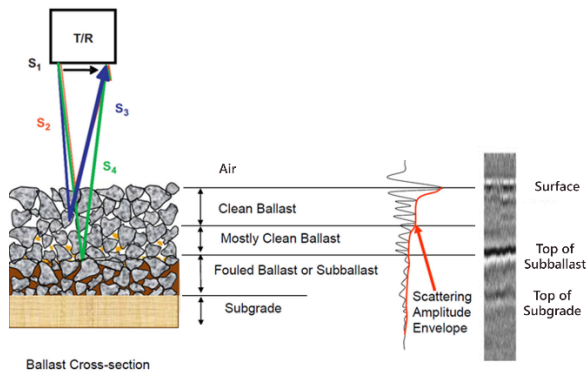

Figure 4. Cross section and A-scan from of a single GPR profile of track layers (adopted from De Bold et al. 2015 & Sui et al. 2023)

GPR technology has demonstrated the capability to estimate the thickness of the ballast and subballast layers with a precision of approximately 3 centimeters (Hugenschmidt 2000). Moreover, it offers some proficiency in determining the extent of subsoil penetration into the ballast layer.

A key property that GPR relies on is dielectric permittivity, which primarily governs the speed of electromagnetic wave propagation through a material, as illustrated in Figure 5. This property is influenced by several factors, including material density, water content, and the type of material. Variations in dielectric permittivity within subsurface materials, as depicted in Figure 5, are critical for GPR's ability to detect subsurface interfaces and anomalies.

Understanding material characteristics is fundamental for professionals working with Ground Penetrating Radar (GPR). Variations in dielectric permittivity significantly impact the accuracy and interpretation of GPR data, playing a crucial role in the success and efficacy of subsurface investigations (Narayanan et al. 2004). It has been observed that fouled ballast possesses a higher dielectric constant compared to clean ballast. This increase is attributed to the presence of contaminants such as moisture, fine particles, and organic matter, which augment its capacity to store electrical energy. These impurities reduce the air voids within the ballast and enhance water retention, thereby contributing to a higher overall dielectric constant detectable by GPR (Anbazhagan et al. 2016).

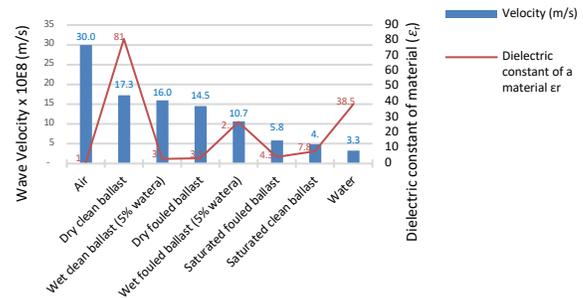

Figure 5. Electromagnetic properties of common materials in ballast layer (De Bold et al. 2015)

Interpreting GPR data involves various formats, each providing unique insights into subsurface features, as illustrated in Figure 6. An A-scan in GPR is a fundamental data form that displays signal strength over time at a single point, useful for identifying individual reflectors and their depths. A B-scan, created by stacking A-scans along a line, provides a 2D cross-sectional view of the subsurface, revealing layers, interfaces, and buried objects. A C-scan, which combines parallel B-scans across an area, offers a 3D plan view of subsurface features, aiding in mapping their extent and distribution. A T-scan, extracting data at a specific depth from B-scans, yields a focused 2D horizontal slice of the subsurface. The selection of a GPR interpretation format is dependent on the specific application and the information required (Kingsuwannaphong et al. 2021).

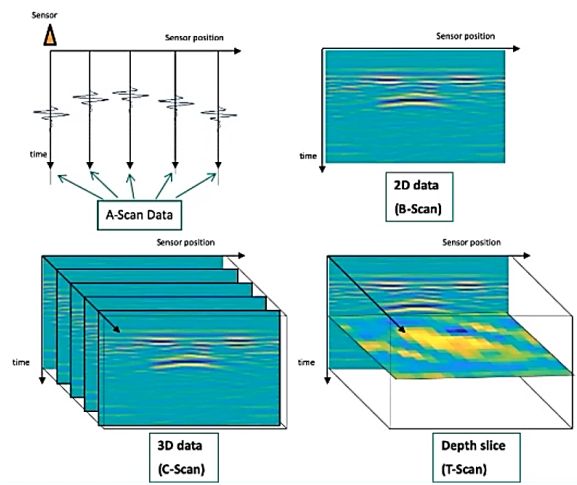

Figure 6. Different forms of GPR interpretation (Kingsuwannaphong et al. 2021)

A series of typical GPR B-scan images depicting various defect types is presented in Figure 7. These images are instrumental in identifying and analyzing subgrade defects, with each type exhibiting distinct features in GPR profiles. Analysis of these images facilitates the recognition and characterization of different defect types, essential for the assessment and maintenance of railway subgrades (Liu et al. 2023).

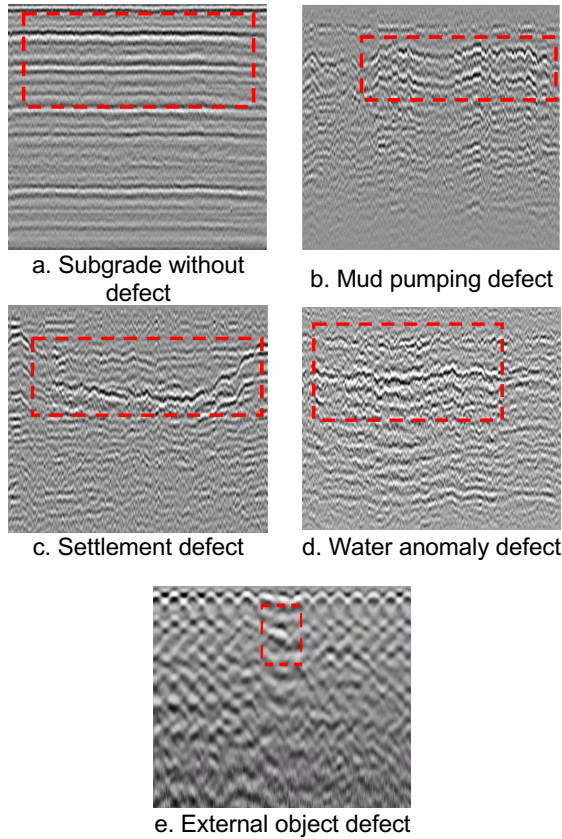

Figure 7. B-scan images of different types of defects in railway subgrades (*Liu, et al. 2023*)

In Figure 7.a, a clear lamellar structure is evident, characterized by a continuous and straight in-phase axis, resulting in uniform reflected energy. In contrast, Figure 7.b shows a disordered, discontinuous, and low-frequency strong reflection pattern, resembling the shape of a mountain peak or a straw hat. Figure 7.c displays a reflection in the settlement radar image, characterized by a notable bending of the in-phase axis and a downward offset in depth. Figure 7.d illustrates signal attenuation, with strong reflected energy at the top surface and the presence of multiple exit waves (Liu et al. 2023). Lastly, Figure 7.e also demonstrates signal attenuation, with strong reflected energy indicating the location of a metallic object, such as a steel pipe.

This paper explores the integration of simulation techniques in ballasted railway tracks and the application of machine learning (ML) models for the precise inspection of various defects, such as cavities, in the process of advanced railway track monitoring using Ground Penetrating Radar (GPR). Key aspects of this technique include identifying the position, depth, and size of defects. Developing a highly accurate machine learning model necessitates a considerable volume of training data, which should encompass a wide range of technical conditions. Acquiring such a dataset from real-world test measurements represents a significant challenge.

The research conducted by Kingsuwannaphong et al. (2021) is pivotal in this context, as it investigates the use of high-accuracy simulation models to generate the required dataset. These simulation models present a substantial advantage by markedly reducing the time required for both the setup and execution of measurements. This efficiency can considerably shorten the data collection timeline, potentially reducing it from years to mere weeks.

3 METHODOLOGY

3.1 Simulation of a Ballasted Railway Track and GPR Test

The use of simulation models is instrumental in calibrating real-world Ground Penetrating Radar (GPR) equipment and validating survey techniques. By comparing actual GPR survey data with simulated models, engineers can more accurately predict the subsurface conditions of railway tracks, thus facilitating more effective maintenance planning (Öztürk et al. 2010). Synthetic modeling of a railway track using GPR involves creating computer-based models that simulate GPR responses under various trackbed conditions. This process is vital for interpreting real-world GPR data with higher accuracy and for identifying and characterizing subsurface features such as ballast condition, moisture content, and the presence of structural anomalies.

Synthetic modeling leads to a better understanding of GPR responses, proving particularly effective in assessing ballast conditions, which exhibit distinct responses based on the extent of fouling, largely due to variations in void size (Barrett et al., 2019). While much of the literature assumes uniform ballast breakdown throughout the ballast layer, it is recognized that ballast contamination often exhibits significant stratification. For instance, in cases where there is a transition between clean ballast and highly fouled ballast, particularly when fouling is stratified, a distinct reflection indicating a boundary layer can be observed in a radargram. This reflection is primarily due to a shift in the effective dielectric constant. However, when this contrast is minimal or the transition from clean to fouled ballast is gradual, the boundary may be less discernible, appearing in the radargram as a change in the degree of scattering (Couchman et al. 2023).

Research involving the creation of synthetic radargrams aims to qualitatively mimic the complex radargrams typically acquired during real surveys. These synthetic radargrams help explore how various trackbed conditions influence GPR responses. To generate these, an open-source electromagnetic simulation software, gprMax, was employed by researchers such as Warren et al. (2016), Couchman et al. (2023), Chomdee et al. (2021), Liu et al. (2020), and Scanlan et al. (2018).

In these studies, gprMax was used to produce radargrams based on trackbed models with predetermined electrical properties. For more complex materials like ballast, external generation was used, with geometries and electrical characteristics imported into gprMax. The ballast layers were generated to reflect realistic particle size distributions (PSDs) and void space ratios, varying according to the PSD used.

Earlier attempts to create ballast models for both mechanical and electromagnetic simulations have employed various approaches. Some electromagnetic models used simplistic 2D particle shapes, such as irregular polygons, without a specific PSD. Others, like Benedetto et al. (2020), incorporated PSD but simplified particle shapes to circles. Striving for more realistic particle shapes, Kingsuwannaphong et al. (2021) used Blender, an open-source graphics software, with a rock generator plugin to create 3D ballast particles that closely resemble real gravel. This approach also incorporated a fouled ballast model by using an effective medium approach within the void spaces (Figure 8). The particle sizes used in their study closely matched industry-standard clean ballast (31.5–63 mm) as defined by Boler et al. (2014) and maintained aspect ratios in accordance with AREMA No. 24 and EN13450 standards. Moreover, Kingsuwannaphong et al. (2021) conducted real GPR measurements with the same ballast size and layer thicknesses for model verification.

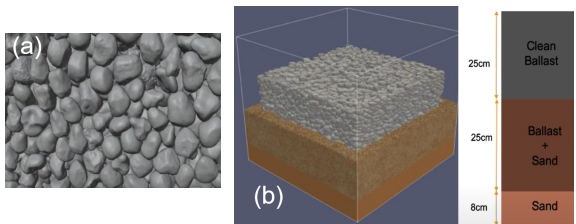

Figure 8. (a). Gravel model and (b). Simulated model of fouled ballast (FDB) (Kingsuwannaphong et al. 2021)

After simulating the GPR test in gprMAX, a comparison of the GPR results of the simulated gravel model with actual measurements shows a high degree of similarity, as expected.

3.2 Application of Machine Learning (ML) Techniques

Machine learning techniques significantly enhance the efficiency and reliability of GPR testing in railway track inspection by automating and improving data analysis. ML algorithms refine GPR data processing, enhancing clarity and reducing noise, thus facilitating better interpretation of subsurface conditions (Kahil et al. 2023). The evolution of GPR-based subgrade defect identification has transitioned from relying on manually designed features to employing ML for classification. For instance, a Support Vector Machine (SVM) approach has been utilized to classify railway ballast types by analyzing amplitude spectral characteristics at frequency inflection points. Another study extracted two-dimensional signal features, such as energy, variance, and histogram statistical features, to develop an SVM-based model for recognizing subgrade defects.

A method involving the division of radargrams into blocks for feature extraction, followed by the use of Fuzzy C-means and Generalized Regression Neural Network algorithms, resulted in high classification accuracies for different defect types (Liu et al. 2023).

Deep learning (DL), a significant advancement in ML (Shomal Zadeh et al. 2023), has been particularly effective in analyzing GPR data for infrastructure health monitoring. Convolutional Neural Networks (CNNs), a DL technique, can be trained to recognize patterns associated with defects in GPR images, thereby swiftly identifying potential issues in railway infrastructure. CNNs are proficient in learning data structure information and dependencies, which enhances target recognition and classification. Besaw et al. (2015) demonstrated high-precision subsurface anomaly identification using CNNs to classify hazardous explosives from GPR data.

Other studies have employed advanced neural network models like YOLO (You Only Look Once) and Cascade R-CNN for detecting mud-pumping defects and subsidence with considerable accuracy (Liu et al. 2023).

YOLO models use a single CNN to simultaneously predict multiple bounding boxes and class probabilities for those boxes. It is known for its speed and accuracy, making it suitable for applications that require real-time pattern recognition images (Li et al et al. 2020). Cascade R-CNN is also an advanced pattern detection framework that improves upon the traditional R-CNN by using a multi-stage approach. While CNNs excel in feature extraction, they have limitations in processing time-series GPR data. This gap is filled by Recurrent Neural Networks (RNNs), capable of learning time-series correlations and dependencies in the data (Tong et al. 2020).

Combining CNN and RNN structures for feature extraction has proven effective, as demonstrated in human re-identification research. The Convolutional Gated Recurrent Neural Network (CGRNN), which combines CNNs for feature extraction with bidirectional gated recurrent units (BGRUs) to learn long-term patterns, has shown enhanced recognition accuracy (Xu et al. 2017).

Liu et al. (2023) combined CNN and RNN for GPR detection of subgrade diseases and anomalies. Their study focused on applying deep neural networks (DNN) for the detection of subgrade defects using raw GPR data. They introduced an innovative DNN model that utilized a multi-layered one-dimensional CNN to autonomously extract feature functions from signal channel waveforms, resulting in a model with fewer parameters and faster processing times compared to previous DNN models using two-dimensional CNNs. Additionally, they incorporated a multi-layer RNN to handle multiple channel features from the CNN, aligning the model more closely with B-scan data. The design of their CNN-RNN (CRNN) model was outlined, and its performance was showcased using manually labeled B-scan data. Their findings indicated that the CRNN achieved comparable accuracy to Faster R-CNN, albeit at a higher speed and with a smaller model size. Faster R-CNN is an advanced object detection neural network that enhances detection speed and accuracy by utilizing a Region Proposal Network (RPN) to identify object locations. This network takes advantage of the strengths of both CNN and RNN, capturing complex patterns and dependencies in the GPR data, making it a powerful tool for detecting railway subgrade conditions and evaluating the presence of defects.

## 4 RESULTS AND DISCUSSION

The work of Kingsuwannaphong et al. (2021) highlighted the difficulty of visually detecting fouling in ballast due to the lack of clear distinctions between Fouled Ballast Data (FDB) and Clean Ballast Data (CDB). This finding underscores the importance of effective detection techniques and precise simulation models in this domain. In their study, a rough estimate of the subgrade interface (sand) location is marked by a red dashed line at approximately 9ns, while the boundary between clean and fouled ballast is denoted by a yellow dashed line around 6ns. Despite efforts to accurately replicate the simulation model, the position of this boundary in the measurements appears slightly shallower (Figure 9). This discrepancy could be due to minor differences in the overall permittivity of the stones compared to those used in the simulation.

These findings indicate that the complex scattering effects caused by ballast present challenges in establishing its visual boundaries, thereby emphasizing the need for the application of advanced machine learning (ML) methods.

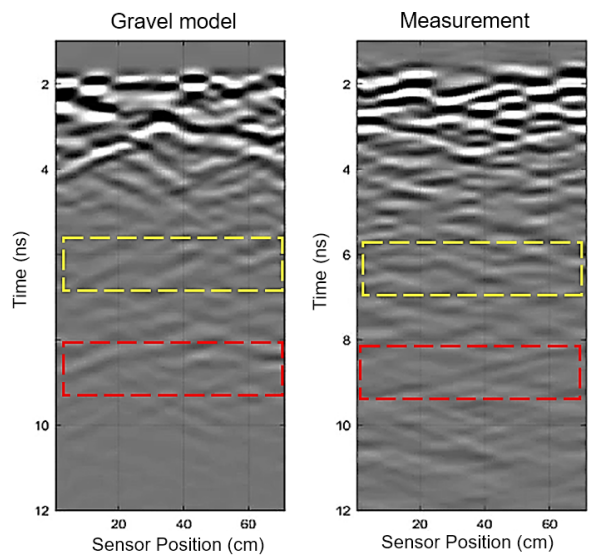

Figure 9. B-Scan comparison between simulated (left) and measured (right) models (Kingsuwannaphong et al. 2021)

Liu et al. (2023) demonstrated through experimental results and a literature survey that four network models – CNN, RNN, CRNN, and Faster R-CNN – each have their respective advantages and disadvantages. The CNN is proficient in extracting deep features from GPR waveforms and reduces storage needs but lacks the capability to process time-series data effectively. The RNN, with its chained architecture, considers both current input and the output of its predecessor network but struggles with gradient issues and nonlinear activation functions. The CRNN model offers the advantages of lightweight, fast inference, and simplicity in handling GPR data, yet it faces limitations in terms of sample size, data imbalance, and lower accuracy. On the other hand, Faster R-CNN delivers high recognition accuracy and has readily available source code, but its model training process is relatively slow. Each model has its specific niche, with its suitability depending on particular requirements and constraints in different scenarios.

## 5 CONCLUSIONS

The innovative application of Ground Penetrating Radar (GPR) in railway track inspection, particularly through the use of simulated models and machine learning (ML), marks a significant advancement in railway infrastructure maintenance and safety. The process of simulating a ballasted railway track and conducting GPR tests, as outlined in this research, underscores the potential of synthetic modeling in accurately predicting the subsurface conditions of tracks, which is crucial for effective maintenance planning. The employment of synthetic modeling, as demonstrated in this study, is instrumental in creating computer-based models that simulate GPR responses under various trackbed conditions, thereby aiding in the precise identification and characterization of subsurface features such as ballast condition, moisture content, and structural anomalies. A notable aspect of this approach is the use of gprMax, an open-source electromagnetic simulation software. gprMax has been pivotal in generating synthetic radargrams that closely mimic real-world scenarios. These radargrams, created with predefined electrical properties and realistic particle size distributions, provide a robust platform for understanding how various trackbed conditions influence GPR responses. The ability of gprMax to produce detailed and accurate simulations has proven invaluable, particularly in providing data for training ML models. This capability is crucial because a significant volume of high-quality data is essential for developing accurate and efficient ML models for railway track inspection. The comparative analysis of GPR results from the gprMax-simulated gravel model with actual measurements further validates the effectiveness of this approach. The close similarity between the simulated and real-world data not only confirms the reliability of the simulation models but also highlights the potential of gprMax in offering a practical and efficient solution for data generation in ML applications.

The Convolutional Recurrent Neural Network (CRNN) represents a groundbreaking model in Ground Penetrating Radar (GPR) anomaly detection, integrating the strengths of Convolutional Neural Network (CNN) and Recurrent Neural Network (RNN) architectures. This model is distinctively designed to process both longitudinal and transverse aspects of GPR data efficiently, obviating the need for transforming data into images. Such a design results in a model that is not only compact but also highly efficient, exhibiting faster processing capabilities than object detection models like Faster R-CNN.

Although the CRNN generally performs on par with its counterparts, it tends to identify more minor anomalies at the boundaries of larger anomalies and occasionally faces challenges in accurately determining starting positions of anomalies. In practical applications, this lightweight model is advantageous as it conserves computational resources by directly processing raw GPR data. This direct approach eliminates the need for image conversion or extensive post-processing for anomaly localization, offering potential for even faster processing with adequate training.

However, the model's primary limitation lies in its focus on anomaly detection; it requires additional data to enhance its classification accuracy, particularly when dealing with complex, real-world data scenarios. Therefore, while the CRNN model marks a significant advancement in GPR data analysis, further refinement and training with diverse datasets are necessary to fully realize its potential in practical applications.

REFERENCES

Al-Qadi, I. L., Xie, W, Roberts, R., (2008). Time-Frequency Approach for Ground Penetrating Radar Data Analysis to Assess Railroad Ballast Condition. Research in Non-destructive Evaluation, 19(4):219-237. DOI: 10.1080/09349840802015107.

Alexander, P.E. (2012). Rail Transportation Energy Efficiency-Oriented Technologies. 787-791. DOI: 10.1115/JRC2012-74130.

Anbazhagan, P., Dixit, P. S. N., & Bharatha, T. P. (2016). Identification of type and degree of railway ballast fouling using ground coupled GPR antennas. Journal of Applied Geophysics, 126, 183–190. DOI: 10.1016/j.jappgeo.2016.01.018.

Annan, A. P. (2009). Electromagnetic principles of ground penetrating radar. Ground penetrating radar: theory and applications, 1, 1-37.

Barrett, B., Day, H., Gascoyne, J. & Eriksen, A. (2019) Understanding the capabilities of GPR for the measurement of ballast fouling conditions. Journal of Applied Geophysics, 169, 183–198.

Besaw, L. E., & Stimac, P. J. (2015). Deep convolutional neural networks for classifying GPR B-scans. In S. S. Bishop & J. C. Isaacs (Eds.), SPIE Proceedings. SPIE. https://doi.org/10.1117/12.2176250.

Boler, H., Qian, Y., & Tutumluer, E. (2014). Influence of Size and Shape Properties of Railroad Ballast on Aggregate Packing: Statistical Analysis. Transportation Research Record, 2448(1), 94-104. https://doi.org/10.3141/2448-12.

Casas, A., Pinto, V., Rivero, L., (2009). Fundamental of ground penetrating radar in environmental and engineering applications. Annals of Geophysics, 43(6):1091-1103. DOI: 10.4401/AG-3689.

Cheng, Y., Li, T., Du, C., Liu, J., Zhang, X., & Zhang, W. (2022). Study on grading evaluation of moisture content in subgrade bed of high-speed railway based on ground penetrating radar technology. Second International Conference on Testing Technology and Automation Engineering (TTAE 2022). https://doi.org/10.1117/12.2660578.

Chomdee, P., Boonpoonga, A., & Lertwiriyaprapa, T. (2021). Study on the detection of object buried under railway by using clutter removal technique. 2021 Research, Invention, and Innovation Congress: Innovation Electricals and Electronics (RI2C), Bangkok, Thailand, DOI: 10.1109/RI2C51727.2021.9559774.

Couchman, M.J., Barrett, B., & Eriksen, A. (2023). Synthetic modelling of railway trackbed for improved understanding of ground penetrating radar responses due to varying conditions. Near Surface Geophysics. DOI:10.1002/nsg.12272.

De Bold, R., O'Connor, G., Morrissey, J.P., Forde, M.C. (2015). Benchmarking large scale GPR experiments on railway ballast, Journal of Construction and Building Materials, Volume 92, Pages 31-42, ISSN 0950-0618, https://doi.org/10.1016/j.conbuildmat.2014.09.036.

Esmaeili, M., Mosayebi, S. A., & Kooban, F. (2014). Effect of Rail Corrugation on the Amount of Train Induced Vibrations Near a Historical Building. Journal of Advances in Railway Engineering, (Vol. 2, No. 2, pp. 73-84).

Feld, R., (2017). Advantages of Electromagnetic Interferometry Applied to Ground-Penetrating Radar: Non-Destructive Inspection and Characterization of the Subsurface Without Transmitting Anything. DOI: 10.4233/UUID:384BF6BE-42DF-4FBA-BBA0-0648C7A52E24.

Hugenschmidt, J. (2000). Railway track inspection using GPR. Journal of Applied Geophysics, 43(2-4), 147-155.

Kahil, N. S., Tempe, V., Yeferni, A., Calon, N., Benkhelfallah, Z., Annag, I., & Mbongo, G. (2023). Automatic analysis of railway ground penetrating radar: Using signal processing and machine learning approaches to assess railroad track substructure. Transportation Research Procedia, 72, 3008–3015. https://doi.org/10.1016/j.trpro.2023.11.848.

Kingsuwannaphong, T., Bräu, C., Rümmler, S., Rial, F., & Heberling, D. (2021). Realistic railway ballast FDTD simulations for ground penetrating radar railway track inspection. 11th International Workshop on Advanced Ground Penetrating Radar (IWAGPR).

Liu, H., Wang, Sh. J., Jing, G., Yu, Z., Guo, Y., (2023). Combined CNN and RNN Neural Networks for GPR Detection of Railway Subgrade Diseases. Sensors, 23(12):5383-5383. DOI: 10.3390/s23125383.

Liu, S., Lu, Q., Li, H., & Wang, Y. (2020). Estimation of moisture content in railway subgrade by ground penetrating radar. Remote Sensing, 12(18), 2912. https://doi.org/10.3390/rs12182912.

Li, Y., Zhen, Z., Luo, Y., & Zhi, Q. (2020). Real-Time Pattern-Recognition of GPR Images with YOLO v3 Implemented by Tensorflow. Sensors, 20(22), 6476. https://doi.org/10.3390/s20226476

Mostofinejad, D., Khademolmomenin, M., & Tayebani, B. (2021). Evaluating durability parameters of concrete containing limestone powder and slag under bacterial remediation. Journal of Building Engineering, 40, 102312. https://doi.org/10.1016/j.jobe.2021.102312

Narayanan, R. M., Jakub, J. W., Li, D., & Elias, S. E. G. (2004). Railroad track modulus estimation using ground penetrating radar measurements. NDT & E International, 37(2), 141–151. DOI:10.1016/j.ndteint.2003.05.003.

Nurmikolu, A., & Guthrie, W. S. (2013). Factors affecting the performance of railway track substructures in seasonally cold climates. ISCORD 2013. https://doi.org/10.1061/9780784412978.063.

O'Neill, K., (1999). Radar sensing of thin surface layers and near-surface buried objects. IEEE Transactions on Geoscience and Remote Sensing, 38(1):480-495. DOI: 10.1109/36.823943.

Öztürk, C., Drahor, M. G., (2010). Synthetic GPR modelling studies on shallow geological properties and its comparison with the real data. 1-4. DOI: 10.1109/ICGPR.2010.5550215.

Scanlan, K. M., Hendry, M. T., Martin, C. D., & Schmitt, D. R. (2018). Evaluating the sensitivity of low-frequency ground-penetrating radar attributes to estimate ballast fines in the presence of variable track foundations through simulation. Proceedings of the Institution of Mechanical Engineers, Part F: Journal of Rail and Rapid Transit, 232(4), 1168–1181. https://doi.org/10.1177/0954409717710408.

Shomal Zadeh, S., Aalipour Birgani, S., Khorshidi, M., & Kooban, F. (2023). Concrete Surface Crack Detection with Convolutional-based Deep Learning Models. International Journal of Novel Research in Civil Structural and Earth Sciences, 10, 25–35. https://doi.org/10.5281/zenodo.10061654.

Solla, M., Vega, P.G., Fontul, S., (2021). A review of GPR application on transport infrastructures: Troubleshooting and best practices. Remote Sensing, 13(4):672-. DOI: 10.3390/RS13040672.

Sui, X., Leng, Z., Wang, S. (2023) Machine learning-based detection of transportation infrastructure internal defects using ground-penetrating radar: a state-of-the-art review, Journal of Intelligent Transportation Infrastructure, Volume 2, liad004, https://doi.org/10.1093/iti/liad004.

Sysyn, M., Nabochenko, O., & Kovalchuk, V. (2020). Experimental investigation of the dynamic behavior of railway track with Sleeper Voids. Railway Engineering Science, 28(3), 290–304. https://doi.org/10.1007/s40534-020-00217-8.

Tong, Z., Gao, J., & Yuan, D. (2020). Advances of deep learning applications in ground-penetrating radar: A survey. Construction and Building Materials, 258(120371), 120371. https://doi.org/10.1016/j.conbuildmat.2020.120371.

The American Railway Engineering and Maintenance-of-way Association: AREMA No. 24 Standards.
The European standards: EN13450-2013, Aggregates for railway ballast Grading D to F.

Travassos, X. L., Pantoja, M. F., (2017). Ground Penetrating Radar. 987-1023. DOI: 10.1007/978-3-319-30050-4_9-1.

Warren, C., Giannopoulos, A., & Giannakis, I. (2016). gprMax: Open-source software to simulate electromagnetic wave propagation for Ground Penetrating Radar, Computer Physics Communications, 209, 163-170, 10.1016/j.cpc.2016.08.020.

Xu, Y., Kong, Q., Huang, Q., Wang, W., Plumbley, M.D. (2017). Convolutional gated recurrent neural network incorporating spatial features for audio tagging. In Proceedings of the 2017 International Joint Conference on Neural Networks (IJCNN), Anchorage, AK, USA, 14–19 May 2017; pp. 3461–3466.